\documentclass[sigconf]{acmart}

\AtBeginDocument{%
  }

\copyrightyear{2025}
\acmYear{2025}
\acmConference[FAccT '25]{The 2025 ACM Conference on Fairness, Accountability, and Transparency}{June 23--26, 2025}{Athens, Greece}
\acmBooktitle{The 2025 ACM Conference on Fairness, Accountability, and Transparency (FAccT '25), June 23--26, 2025, Athens, Greece}\acmDOI{10.1145/3715275.3732020}
\acmISBN{979-8-4007-1482-5/2025/06}

\author{Henrik Nolte}
\authornote{Both authors contributed equally to this research.}
\affiliation{%
  \institution{University of Tübingen}
   \city{Tübingen}
  \country{Germany}
  }
\email{henrik.nolte@uni-tuebingen.de}

\author{Miriam Rateike}
\authornotemark[1]
\affiliation{%
  \institution{IBM Research Africa \\  Saarland University}
     \city{Nairobi}
  \country{Kenya}}
\email{miriam.rateike@ibm.com}

\author{Michèle Finck}
\affiliation{%
  \institution{CSZ Institute for Artificial Intelligence and Law, University of Tübingen}
     \city{Tübingen}
  \country{Germany}
    }

\usepackage{multirow} 
\usepackage{subcaption} 

\usepackage{amsthm} 
\usepackage{amsfonts} 
\usepackage{hyperref} 
\usepackage{url} 
\usepackage{graphicx}  
\usepackage{pifont} 
\newcommand{\cmark}{\ding{51}}%
\newcommand{\xmark}{\ding{55}}%

\everypar{\looseness=-1}
\newcommand{\EUAIAct}{AIA} 
\newcommand{\EW}[1]{Rec.~(#1)}
\newcommand{\Rec}[1]{Rec.~(#1)}
\newcommand{\EWx}{Rec.}
\newcommand{\Art}[1]{Art.~#1~\EUAIAct}
\newcommand{\Artx}{Art.}

\newcommand{\HRAIS}{HRAIS}

\newcommand{\GPAIM}{GPAIM}
\newcommand{\GPAIMS}{GPAIMs}

\newcommand{\GPAIMSSR}{GPAIMs with systemic risk}

\begin{document}

\title{Robustness and Cybersecurity in the EU Artificial Intelligence Act}

\begin{abstract}
The EU Artificial Intelligence Act (AIA) establishes different legal principles
for different types of AI systems.
While prior work has sought to clarify 
some of these principles, 
little attention has been paid to robustness and cybersecurity.
This paper aims to fill this gap.
We identify legal challenges and shortcomings in provisions related to robustness and cybersecurity for high-risk AI systems (\Art{15}) and general-purpose AI models (\Art{55}).
We show that robustness and cybersecurity demand resilience against performance disruptions.
Furthermore, we assess potential challenges in implementing these provisions in light of recent advancements in the machine learning (ML) literature.
Our analysis informs efforts to develop harmonized standards, guidelines by the European Commission, as well as benchmarks and measurement methodologies under \Art{15(2)}. 
With this, we seek to bridge the gap between legal terminology and ML research, fostering a better alignment between research and implementation efforts.
\end{abstract}



\maketitle

\section{Introduction}\label{sec:intro}
The European Union (EU) recently adopted the Artificial Intelligence Act (\EUAIAct)\footnote{EU Regulation 2024/1689, 12.7.2024.} which creates a legal framework for the development, deployment, and use of ``human-centred and trustworthy artificial intelligence (AI)'' (\Art{1}).
The \EUAIAct\ outlines 
desirable ``ethical principles'' of AI systems (\EW{27}) and, i.a., imposes some of these as legally binding requirements for high-risk AI systems (\HRAIS), e.g, AI systems intended to be used to take university admission decisions, or to evaluate individuals' creditworthiness, and for general-purpose AI models (\GPAIMS), e.g., multimodal large language models.
While the \EUAIAct\ is recognized as being one of the first legally binding regulatory frameworks for AI~\citep{Chee_Hummel_2024}, it has faced criticism for its imprecise and incoherent terminology~\citep{laux2024trustworthy, bomhard2024AIAct}, which will complicate its practical implementation. 
Previous work has examined the \EUAIAct\ and its legislative history to clarify terms like explainability~\citep{bordt2022post, vitali2022survey, pavlidis2024unlocking} and fairness~\citep{deck2024implications}. So far, little attention has been paid to robustness and cybersecurity.
Only \emph{AI systems} classified as high-risk (\HRAIS) must meet the robustness and cybersecurity requirements set out in \Art{15}. This paper thus focuses on requirements for \HRAIS. To provide a clearer understanding, we compare these requirements with requirements for specific \emph{AI models}, namely for \GPAIMSSR, in \Art{55}.

Technical solutions to ensure the robustness and cybersecurity of AI systems are often developed within the ML domain. 
Therefore, it is essential to inform ML research about the legal requirements to ensure compliance with the \EUAIAct.
However, the vagueness of requirements for cybersecurity and robustness under the \EUAIAct\ makes it challenging to inform ML practitioners about the specific legal requirements to further the development of solutions that can ensure compliance with the AIA.
A common understanding between technical and legal
domains can be facilitated through technical standards. 
While the \EUAIAct\ sets out general rules, technical standards specify these rules in detail. 
Standards are technical specifications designed to provide voluntary technical or quality specifications for current or future products, processes
or services.\footnote{\Artx~1, 2(1) EU Regulation 1025/2012, OJ L 316, 14.11.2012.}
They prescribe technical requirements, 
including characteristics such as quality or performance levels, terminology, and test methods.\footnote{\Artx\ 2(4)(a) and (c) ibid.}
Standards have long been integral to EU product legislation under the New Legislative Framework, upon which the AIA is built~\citep{gorywoda2009new}.
If approved by the EU Commission, technical standards become harmonized technical standards, which grants a presumption of conformity to products or processes that adhere to them. Consequently, compliance with these standards is deemed to fulfil the requirements of the \EUAIAct, thereby incentivising providers to adopt them  (\Art{40}).
The development of harmonized technical standards for the \EUAIAct\ has been initiated by the EU Commission
and is expected to be completed in the next few years.
In addition, the EU Commission is also tasked with developing additional guidelines on the practical implementation on the application of \Art{15} (\Art{96(1)(a)}).

In this paper, we make the following contributions:
\begin{itemize}
    \item We analyse and explain the legal requirements related to robustness and cybersecurity in the \EUAIAct, identify related shortcomings, and offer possible solutions for
    some of these shortcomings. 
    \item 
    We evaluate these findings in relation to their practical implementability.
    This aims to inform the standardization process, the development of guidelines by the EU Commission, as well as the benchmark and measurement methodologies referred to in \Art{15(2)}.
    \item We connect the legal requirements for robustness and cybersecurity to ML terminology, aiming to inform ML research and ensure that technical solutions are conducive to legal compliance.
\end{itemize}

Our research employs a doctrinal analysis and as such involves a systematic examination and interpretation of provisions of the AIA, starting with \Art{15} and extending to other provisions within the AIA. We refer to recitals in our analysis as they are likely a starting point of any future judicial engagement with these legal provisions. Indeed, whereas recitals are not strictly legally binding they have been highly authoritative in Court of Justice of the European Union (CJEU) case law. While our analysis is primarily grounded in legal scholarship, we map our findings to machine learning (ML) literature to bridge the gap between the legal and ML domains. This approach aims to inform a broader audience about the regulatory challenges and opportunities at the intersection of these fields. Our analysis follows the principle that law should be interpreted autonomously, recognizing that legal terminology does not necessarily align with technical terminology.

This paper is structured as follows: Section~\ref{sec:background_robustness} provides a short background on robustness and cybersecurity in the ML literature.
Section~\ref{sec:background_aia} provides an introduction to the \EUAIAct\ and \Art{15}.
Section~\ref{sec:legal_challenges} analyses the requirements outlined in \Art{15} for \HRAIS, addressing both general challenges pertinent to robustness and cybersecurity, as well as specific issues related to each requirement. 
Section~\ref{sec:gpai} examines the requirements in \Art{55} relevant to \GPAIMSSR. 
Section~\ref{sec:discussion_outlook} concludes with a summary and recommendations for future research.

\section{An ML Perspective on Robustness and Cybersecurity}
\label{sec:background_robustness}
ML research on robustness focuses on mitigating undesired 
changes in model outputs when deploying models in real world scenarios~\cite{schwinn2022improving}. This issue is explored across different applications such as computer vision~\cite{drenkow2021systematic, taori2020measuring, dong2020benchmarking} and natural language processing~\cite{la2022king, chang2021robustness}.
Unintended changes in model outputs can occur due to adversarial or non-adversarial factors affecting the ML model, its input (test) data, or its training data~\cite{cheng2024towards, tocchetti2022ai}.

\begin{figure*}
    \centering
    \begin{minipage}[t]{0.51\textwidth}
        \centering
        \includegraphics[width=\textwidth]{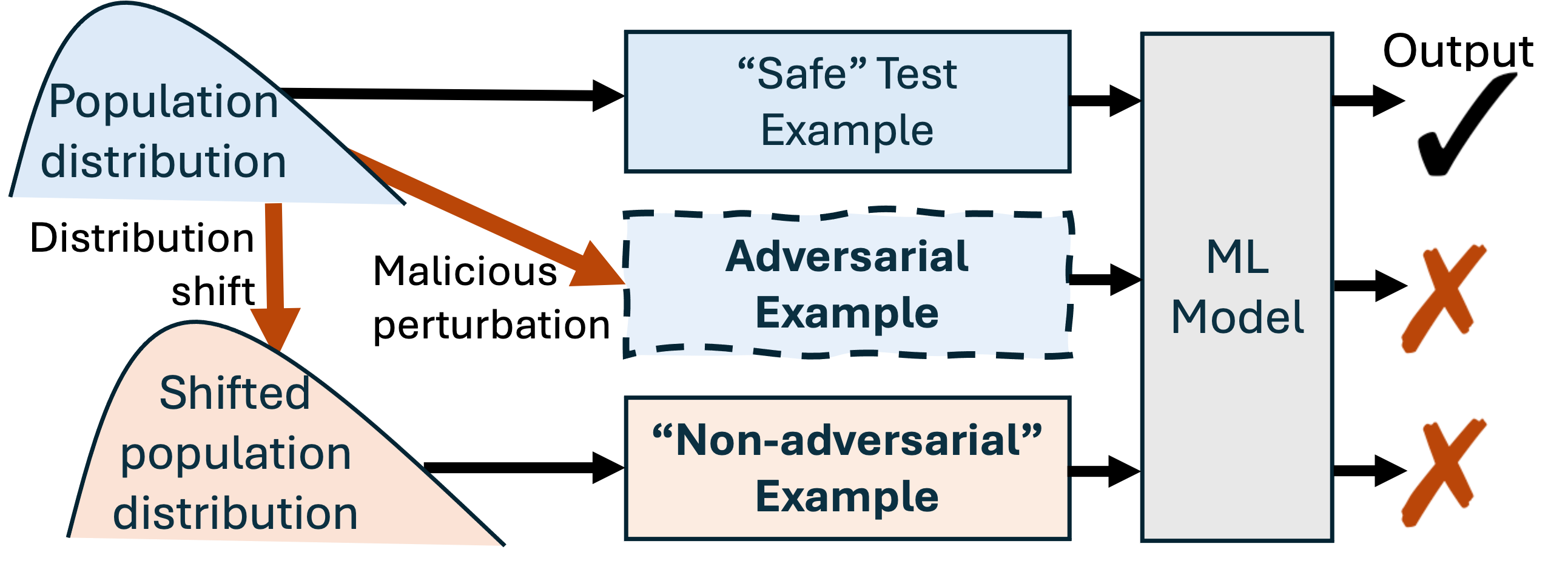}
        \caption{Robustness problems. Outputs may be as expected (\cmark) with ``safe'' test data from the original distribution; unintended changes (\xmark) can occur with adversarial or non-adversarial (shifted) inputs.}
        \Description[A flowchart illustrating how different types of inputs influence a machine learning (ML) model's performance.]{A test example sampled from the original population distribution is labelled as a safe example and, when passed to the ML model, results in a correct output. If this example is maliciously perturbed, it becomes an adversarial example, which causes the ML model to produce an incorrect output. Separately, a distribution shift on the population distribution leads to a shifted population distribution. Sampling from this shifted population distribution produces a non-adversarial example, which, when passed to the ML model, also leads to an incorrect prediction.}
        \label{fig:robustness}
    \end{minipage}
    \hfill
    \begin{minipage}[t]{0.47\textwidth}
        \centering
        \includegraphics[width=0.8\textwidth]{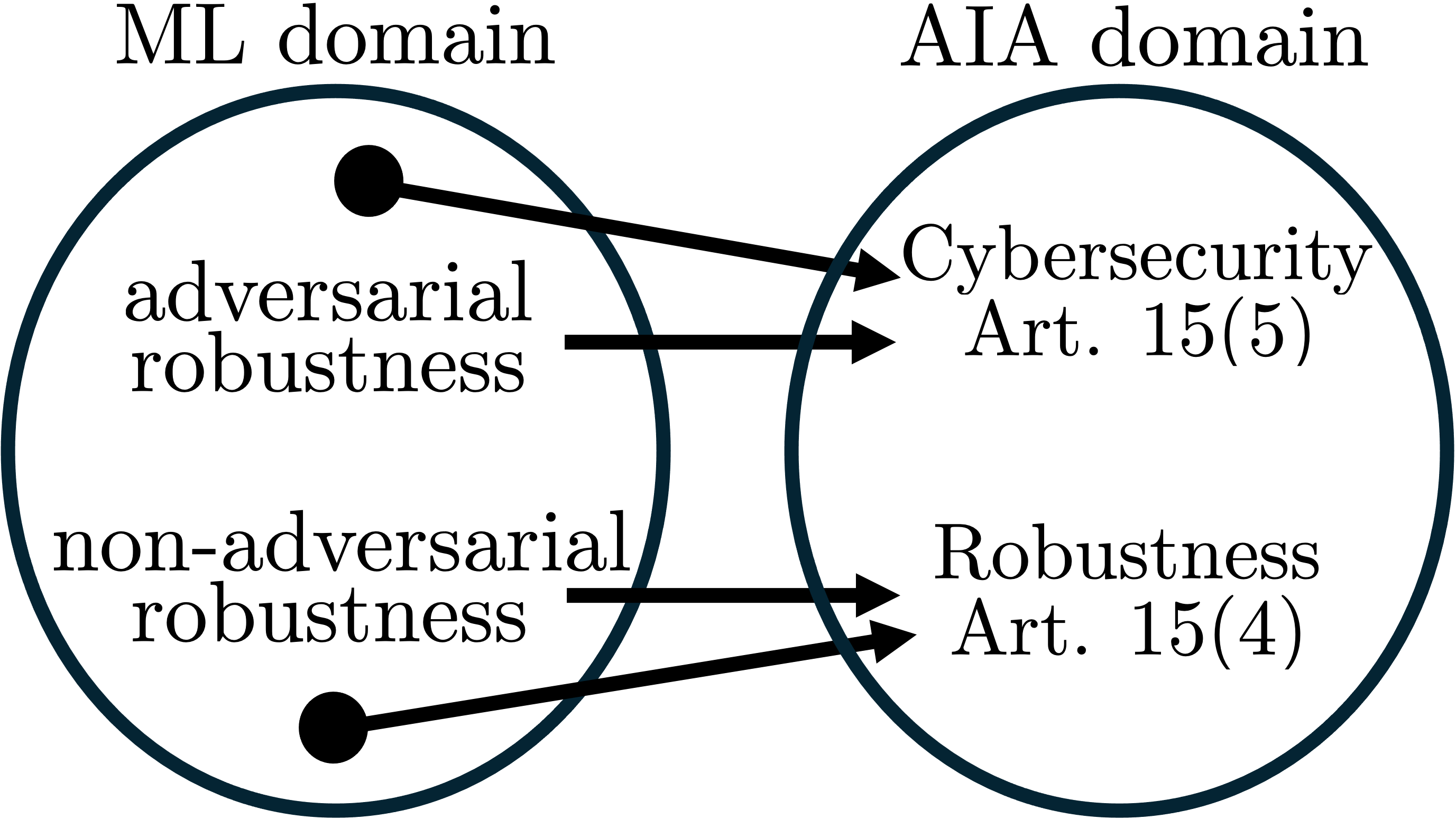}
        \caption{Technical solutions to cybersecurity can be found, i.a., in ML research on \emph{adversarial robustness}; technical solutions to robustness can be found, i.a., in the ML research on \emph{non-adversarial robustness}.}
        \Description[Overview of mapping between the Machine Learning (ML) domain and the Artificial Intelligence Act (AIA) domain.]{The image shows two non-overlapping circles, one representing the Machine Learning (ML) domain and one the Artificial Intelligence Act (AIA) domain. The ML domain includes adversarial robustness and non-adversarial robustness. Arrows connect these concepts to the AIA domain. Adversarial robustness is linked to Cybersecurity, Art. 15(5) AIA. Non-adversarial robustness connects to Robustness, Ar. 15(4) AIA.}
        \label{fig:mapping}
    \end{minipage}
\end{figure*}

Perturbations of input (test) data often present a difficult challenge (see Figure~\ref{fig:robustness}).
While a model's output may be as expected when using ``safe'' 
test data from the original population distribution, unintended changes can occur when perturbed examples are
provided as input to the ML model.

\emph{Non-adversarial} (or \emph{natural}) \emph{robustness} often addresses changes in ML model outputs due to \emph{distribution shifts} in input data~\citep{tocchetti2022ai, gojic2023non}.
These changes occur when the distribution from which the test data is sampled differs from that of the training data~\citep{drenkow2021systematic, taori2020measuring}.
For instance, alterations in data collection methods, such as upgrading to a new X-ray machine, can modify the format or presentation of images~\cite{glocker2019machine, castro2020causality}. Importantly, distribution shifts can also result from \emph{feedback loops}, where the ML model's outputs influence the data distribution, creating a cycle from the model's output back to its input~\cite{d2020fairness, zhang2020fair}.
Such an effect can be found, for example, in movie recommendation systems, where user’s preferences change over time in response to the ML system’s suggestions, thereby influencing future recommendations~\cite{perdomo2020performative}.
Other forms of research on non-adversarial robustness investigate the robustness of ML models to noise, which frequently occurs in real-world data sets~\cite{saez2016evaluating, olmin2022robustness}.

\emph{Adversarial robustness} refers to the study and mitigation of model evasion attacks using adversarial examples. 
These are data samples typically drawn from the original population distribution and then modified by an adversary---often in ways that are difficult or even impossible to detect through human oversight---with the intent of altering a model's output~\citep{szegedy2013intriguing}. 
This phenomenon can occur across various models and data types.
For instance, in the image domain, small pixel perturbations in input images can lead to significant changes in a model's output~\citep{szegedy2013intriguing}.
In a broader sense, adversarial robustness also encompasses the study and mitigation of other forms of adversarial attacks that attempt to extract the model or reconstruct or perturb the training data set~\cite{nicolae2018adversarial, chen2017targeted}.

From a technical standpoint, adversarial robustness is one aspect of \emph{cybersecurity}. 
Research on cybersecurity focuses on developing defences that protect computer systems from attacks compromising their confidentiality, integrity, or availability~\cite{dasgupta2022machine}. 
This encompasses aspects like data storage, information access and modification, and secure data transmission over networks~\cite{sarker2021ai}.
Unlike robustness, cybersecurity is not a stand-alone concept in ML,
but is discussed more broadly as both a tool for ensuring cybersecurity and a potential source of cybersecurity risks.\footnote{E.g., NeurIPS'18 Workshop on Security in ML~\cite{secml2018}, ICML'22 Workshop on ML for Cybersecurity~\cite{ml4cyber}.}
ML algorithms can be employed to detect and mitigate cybersecurity threats~\cite{sarker2021ai}, 
but can also introduce specific vulnerabilities that adversaries may exploit, such as data poisoning or adversarial attacks~\cite{roshanaei2024navigating, rosenberg2021adversarial}.

\section{Background on the \EUAIAct\ and \Art{15}}
\label{sec:background_aia}

\paragraph{\EUAIAct} 
The \EUAIAct\ creates harmonized rules for certain AIA systems in order to incentivize the use of such systems in the EU market and prevent regulatory fragmentation between member states (\EW{1}).
It is formally structured into recitals (\EWx), articles (\Artx), and annexes. Recitals are not strictly legally binding and outline the rationale behind the articles, articles delineate specific binding
obligations, and the annexes provide additional details and 
specifications to support the articles
~\citep{klimas15law}.\footnote{From now on, whenever we cite recitals, we refer to those in the \EUAIAct\ without explicitly indicating it.}
\Art{3(1)} defines an AI system as ``a machine-based system that is designed to operate with varying levels
of autonomy and that may exhibit adaptiveness after deployment, and that, for explicit or
implicit objectives, infers, from the input it receives, how to generate outputs [...] that can influence physical or virtual
environments''. 
These AI systems are regulated differently based on their perceived risk level~\citep{bomhard2024AIAct, sioli2021}: Those posing unacceptable risks, such as social scoring, are prohibited or subject to qualified prohibitions; \HRAIS, such as those used in 
medical devices, are allowed but must comply with certain requirements and undergo pre-assessment; other AI systems are subject only to specific transparency and information obligations. 
Among these categories, only \HRAIS\ must fulfil the robustness and cybersecurity requirements under \Art{15}.
According to \Art{16(a)}, providers of \HRAIS\ must ensure compliance with these requirements. 
%
An AI system is considered a \HRAIS\ if it is either a safety component of a product or a product itself regulated under specific legislation, such as medical devices, machinery, or toys (\Art{6(1)}, Annex I), or if it poses a significant risk of harm to the health, safety, or fundamental rights of individuals in specific areas, such as education, employment, or law enforcement (\Artx\ 6(2) and (3) \EUAIAct, Annex III).
%

In addition to AI systems, the \EUAIAct\ establishes a separate regime of legal requirements in chapter V of the \EUAIAct\ for a very specific type of AI models, namely \GPAIM\ (e.g., multimodal large language models, see also Section~\ref{sec:gpai}).
Note that the AIA differentiates between AI systems and AI models. While \EW{97} specifies that an AI model is an essential component of an AI system, there is no clear legal delineation of the two concepts, as we discuss further in Section 5.1.
GPAI models are AI models that can perform tasks that they were not originally trained for~\citep{gutierrez2023proposal}, such as large language models~\cite{openai2023gpt, team2023gemini}, or large text-to-image models~\cite{ramesh2022hierarchical}. Other types of AI models are not regulated and mentioned in the \EUAIAct.
These are models that are created with a specific objective and can only accomplish tasks they are trained to perform (e.g., translation, classification).

The \EUAIAct\
does not directly define specific technical requirements. Instead, it sets out `essential requirements' that AI systems
must comply with, and which are concretized by so-called technical standards.
The regulatory concept of relying on standardization is a well-established process in EU product legislation and is referred to as the New Legislative Framework~\citep{gorywoda2009new}.
It traditionally involves the participation of stakeholders, such as providers of AI systems.
The European Commission has already issued a standardization request to the European Committee for Standardisation and the European Committee for Electrotechnical Standardisation to develop standards 
for the \EUAIAct\ until 30 April 2025.\footnote{C(2023)3215 - Standardisation request M/593.}
Specifically, Annex I enlists  `robustness' and `cybersecurity' specifications for AI systems as standardisation deliverables to be developed.
Once 
accepted by the European Commission, these technical standards become `harmonized' standards
and compliance with the \EUAIAct\ will be presumed if providers adhere to them
(\Art{40}).

\paragraph{\Art{15}} 
\Art{15(1)} requires that \HRAIS\ ``shall be designed and developed in such a way that they achieve an appropriate level of accuracy, robustness, and cybersecurity, and that they perform consistently in those respects throughout their lifecycle''.
The provision outlines specific product-related requirements for AI systems to ensure they are trustworthy.
\Art{15(4)} mandates that \HRAIS\ exhibit resilience ``regarding errors, faults or inconsistencies that may occur within the system or the environment in which the system operates, in particular due to their interaction with natural persons or other systems''.
Additionally, \Art{15(5)} requires \HRAIS\ to be ``resilient against attempts by unauthorised third parties to alter their use, outputs or performance by exploiting system
vulnerabilities''. 
\Art{15(2)} requires the EU Commission, together with other relevant stakeholders, to encourage the development of benchmarks and measurement methods for assessing accuracy, robustness, and other performance metrics.

\section{The Purpose of \Art{15}}\label{sec:purpose_15aia}
\Art{15(1)} states that high-risk AI systems must achieve an ``appropriate level of accuracy, robustness and cybersecurity'' and must function consistently in this respect throughout their lifecycle. It is one of seven provisions in the AIA that sets specific requirements for high-risk AI systems. 
These requirements aim to ensure product safety and 
a ``consistent and high level of protection of public interests as regards health, safety and fundamental rights'' (\EW{7}). Particularly, \Art{15(1)} outlines specific product-related requirements for AI systems, detailing how they must be ``designed and developed'' to achieve trustworthy AI. 
This is emphasized by the 2019 Ethics Guidelines for Trustworthy AI developed by
the AI Independent High-level Expert Group (AI IHEG) on AI~\citep{aiiheg2019guidelines} appointed by the European Commission which can be seen as the conceptual basis of the \EUAIAct\ (\EW{27}). 
These guidelines state that technical robustness ensures that AI systems ``reliably behave as intended while minimizing unintentional and unexpected harm, and preventing unacceptable harm''.

The impact assessment clarifies that ``is important that these requirements are included in the regulatory framework so that
substandard operators need to improve their procedures''~\citep[Annex 4]{assessment2021commission}.
Therefore, the purpose of the specific requirements set out in \Art{15} is to achieve the overarching objective of ensuring the trustworthiness of AI systems (see \Art{1(1)}) and to advance the cybersecurity agenda of the EU.\footnote{COM/2010/245 final/2, COM/2021/118 final.}

In particular, \Art{15} interacts with numerous other legal requirements that impose direct or indirect cybersecurity-related obligations on products that can qualify as HRAIS, such as the Machinery Regulation, the Medical Device Regulation, and the European Health Data Space. While cybersecurity has become a central concern for lawmakers in the EU, the arising frameworks create a patchwork of legislation that is difficult to disentangle. \Art{15} adds another layer to cybersecurity-related obligations in the EU. 

\section{Requirements for High-Risk AI Systems}\label{sec:legal_challenges}
In this section, we provide an analysis of the overarching challenges of implementing \Art{15} (Section~\ref{sec:general_challenges}), followed by a discussion regarding the robustness requirement in \Art{15(4)} (Section~\ref{sec:challenges_robustness}) and the cybersecurity requirement in \Art{15(5)} (Section~\ref{sec:challenges_cybersecurity}).

\subsection{General Challenges of \Art{15}}\label{sec:general_challenges}

We identify four legal challenges related to \Art{15} that may arise in its practical implementation.
 First, there is no clear delineation of the legal terms of robustness and cybersecurity and its counterparts in ML literature.
 Second, while the \EUAIAct\ mandates compliance for entire AI systems, the ML literature primarily focuses on models, which may pose practical challenges for implementation. 
Third, while accuracy is specified as a requirement in \Art{15}, the provision does not clarify its role in measuring robustness and cybersecurity.
Fourth, the terms 'lifecycle' and 'consistent' performance are not defined, leaving ambiguity about how such performance can be practically ensured.

\paragraph{Legal Requirements.}

The robustness requirement in \Art{15(4)} addresses ``errors, faults, or inconsistencies'' that may inadvertently occur as the system interacts with its real-world environment. 
In contrast, the cybersecurity requirement in \Art{15(5)} targets deliberate attempts ``to alter the use, outputs, or performance'' of an AI system ``by malicious third parties exploiting the system’s vulnerabilities''. 
Both robustness and cybersecurity requirements aim to ensure that \HRAIS\ perform consistently and are resilient against any factors that might compromise this performance. They, however, address different threats to consistent performance: robustness requires protection against unintentional causes, whereas cybersecurity protects against intentional actions.
While robustness is a new term in EU legislation and not explicitly defined in the \EUAIAct, the term cybersecurity has already been defined in the EU Cybersecurity Act (CSA)\footnote{Regulation (EU) 2019/881, OJ L 151, 7.6.2019.}.
Art. 2(1) CSA provides a broad definition of cybersecurity, covering all ``the activities necessary to protect network and information systems, the users of such systems, and other persons affected by cyberthreats''. According to the CSA, a cyber threat is any potential circumstance, event or action that could damage, disrupt or otherwise adversely impact network and information systems, the users of such systems and other persons (Art. 2(8) CSA). Importantly, the CSA does not distinguish between intentional or unintentional cyberthreats as causes of harm. Rather, both are explicitly included in the scope of the CSA (see, e.g., Art. 51(1)(a) and (b) CSA).
However, the \EUAIAct\ artificially splits the CSA's concept of cybersecurity by designating unintentional causes as a matter of robustness and restricting cybersecurity to intentional actions. This creates a conflict when aligning the AIA's requirements with the CSA's definition of cybersecurity that may lead to regulatory ambiguity. Specifically, \Art{42(2)} considers \HRAIS\ with CSA certification or conformity declarations as compliant with cybersecurity requirements in \Art{15}.\footnote{Note that this holds only true in so far as the cybersecurity certificate or statement of conformity or parts thereof cover those requirements in \Art{15}.} This suggests that the CSA definition of cybersecurity applies to the \EUAIAct, even though it inherently covers both types of causes.

\paragraph{Mapping to ML Understanding of Adversarial and Non-adversarial Robustness} We explore how these legal terms could be understood within the ML domain proposing a simple model as an explanatory heuristic (see Figure~\ref{fig:mapping}).
In ML, robustness refers to maintaining consistent model performance in real-world scenarios~\cite{schwinn2022improving}. 
ML research distinguishes between different types of robustness. \emph{Non-adversarial robustness} in ML refers to a model’s ability to maintain performance despite data shifts or noise~\cite{tocchetti2022ai, gojic2023non, saez2016evaluating, olmin2022robustness}. This aligns with the legal term robustness in the \EUAIAct. 
\emph{Adversarial robustness} in ML refers to the model’s resistance to intentional perturbations aimed at altering predictions~\cite{szegedy2013intriguing}. This aspect aligns more with the legal concept of cybersecurity.
The cybersecurity requirement in \Art{15(5)} aims to ensure AI systems' integrity, confidentiality, and availability, protecting them from threats like unauthorized access, adversarial manipulation, data modification, Denial-of-Service attacks, and theft of sensitive information (e.g., model weights).
However, other scenarios within the ML domain may also fall under the relevant legal terms. For example, language model jailbreaks exploit AI vulnerabilities to bypass safety constraints~\cite{wei2024jailbroken, vassilev2024adversarial}. This aligns more closely with the notion of cybersecurity in protecting against misuse of AI systems.

Our findings are supported by an historic analysis of the legislation process.
As outlined in Section~\ref{sec:purpose_15aia}, the \EUAIAct\ builds on the Ethics Guidelines for Trustworthy AI~\cite{aiiheg2019guidelines}.
In the guidelines, the principle of `technical robustness and safety' includes resilience against attacks, but does not mention cybersecurity.   
The White Paper on Artificial Intelligence~\cite{whitepaper}, which elaborates on these guidelines, still lists resilience to attacks against AI systems under ``robustness and accuracy'' without differentiating those terms from cybersecurity.
The first official draft of the \EUAIAct\ by the European Commission\footnote{COM/2021/206 final.} was the first official document to distinguish between these three terms and assigned ``resilience against attacks'' to cybersecurity rather than robustness.
\EW{27}, which refers to the IHGE guidelines, seems to be a remnant of this development process.
It demands under the term `technical robustness' that AI systems should be resilient ``against attempts to alter the use or performance of the AI system'', essentially asking for adversarial robustness.

\paragraph{System vs. Model.}
The \EUAIAct\ regulates AI systems, but not AI models, with the only exception being \GPAIMS. 
Accordingly, the wording of \Art{15} refers to ``systems''.
ML research, 
in contrast, 
often focuses on developing technical solutions for \emph{ML models}.
This raises the practical question of whether solely relying on technical solutions for \emph{ML models} is enough to also ensure the compliance of a \HRAIS\ with \Art{15}---or 
whether and how cybersecurity and robustness should be evaluated at the system level.

\EW{97} 
specifies that an AI model is an essential component of an AI system.\footnote{Although \Rec{97} specifically refers to \GPAIMS, the wording suggests that the statement about the relationship between AI systems and AI models is of a general nature.}
Additional components can include, i.a., user interfaces, sensors, databases, network communication components, or pre- and post-processing mechanisms for model in- and outputs (\EW{97}, \cite{JRC134461}).
All these individual components should contribute to the overall robustness of the AI system, particularly in scenarios where some components may fail.
This is illustrated by \Art{15(4)(ii)}, which
states that robustness may be ensured through technical redundancy solutions, including ``back-up or contingency plans''.
%
Furthermore, \Art{15(5)(iii)} stipulates that the cybersecurity of AI systems shall be achieved through technical solutions that, ``where appropriate'', target training data, pre-trained components, the AI model or its inputs. 
This binding provision suggests that at least these different components of the AI system are required to be assessed individually for their appropriateness in mitigating cybersecurity attacks.
Thus, \Art{15} should not be understood as requiring a single, unified assessment of the requirements. Instead, it must be interpreted as mandating that each component, including one or more ML models, be assessed individually. The assessment of the AI system’s overall performance is then derived from an aggregation of the individual performance results~\cite{kumar2023}. 
This requires an interdisciplinary approach that draws on expertise from fields such as ML, engineering, and human-computer interaction.
To establish a common understanding, it can prove beneficial to formally describe the evaluation process of an entire AI system, including potential challenges, such as interdependencies of technical solutions. 

\paragraph{Role of Accuracy.}

\Art{15(1)} mandates that \HRAIS\ shall ``achieve an
appropriate level of accuracy''.
This is important because trade-offs between different desiderata can exist, such as between robustness and accuracy (see Appendix~\ref{apx:robustness-accuracy-trade-off}).
While accuracy is not defined in the \EUAIAct, Annex~IV~No.~3~\EUAIAct\ states that accuracy is an indicator of the capabilities and performance limits of an AI system.
Accordingly,  accuracy should be measured in at least two ways:
i) separately for ``specific persons or groups of persons on which the system is intended to be used''\footnote{This links to fairness ML literature on diverging error rates for different sensitive groups~\cite{mitchell2021algorithmic, chouldechova2017fairer}.}, and
ii) the overall expected accuracy for the ``intended purpose'' of the AI system. 
In ML, the metric \emph{accuracy} typically describes the overall proportion of correct predictions out of the total number of predictions made~\cite{carvalho2019machine}. 
However, the term can also describe the objective of ``good performance'' of an AI system and, depending on its specific purpose, can also be evaluated using different metrics, such as utility~\cite{corbett2017algorithmic} and f1-score~\cite{sokolova2006beyond}. 
\Art{15(3)} explicitly references  `accuracy and the relevant accuracy metrics', indicating 
that accuracy is understood as an objective that can be measured with various metrics, leaving the choice of the relevant metric to the provider.
The selection of the metric should consider various factors, including the specific purposes of the ML model, dataset-specific circumstances (e.g., imbalanced data) and the particular model type (e.g., classification, regression). Technical standards and guidelines by the EU Commission should clarify how AI systems' accuracy should be measured.

In ML, robustness is often measured using an \emph{accuracy} metric.
Typically, this involves comparing the \emph{accuracy} (or error rates) evaluated on an unperturbed dataset from the original distribution with the accuracy on a perturbed test set (e.g., sampled from the shifted distribution or containing adversarial samples)~\cite{taori2020measuring, hendrycks2021many, goodfellow2014explaining}. 
The smaller the difference between these two \emph{accuracy} results, the better the \emph{robustness}.
The choice of the \emph{accuracy} metric thus has an impact on the measurement of robustness. 
As a result, the ML model may appear more robust under some accuracy metrics than others.
The selection of favourable metrics has been studied in fair ML under the term fairness hacking~\cite{meding2024fairness, simson2024one, black2024d}.
Without entering into the debate, we note that there is an ongoing discussion in the ML literature about the existence and characteristics of a trade-off between \emph{robustness} and \emph{accuracy}. 
While some research showed that enhancing \emph{robustness} leads to a drop in \emph{test accuracy}~\cite{zhang2019theoretically, rade2022reducing, tsipras2018robustness}, others believe that \emph{robustness} and \emph{accuracy} are not conflicting goals and can be achieved simultaneously~\cite{yang2020closer, raghunathan2020understanding}.
Technical standards and guidelines by the EU Commission should provide instructions on how AI system providers should choose an appropriate `accuracy' measure, especially when it is used to assess robustness in subsequent steps.

\paragraph{Consistent Performance Throughout the Lifecycle.}

AI systems must perform ``consistently'' in terms of accuracy, robustness, and cybersecurity ``throughout their lifecycle'' (\Art{15(1)}). 
Performance is the ``ability of an AI system to achieve its intended purpose'' (\Art{3(18)}).
However, i) the term `lifecycle' is not defined, creating ambiguity about whether it differs from the term `lifetime' used in \Art{12(1)} and \EW{71}; ii) the concept of `consistent' performance is unclear, and it is not specified how it should be measured.

First, `lifecycle' and `lifetime' could be understood as synonyms \cite{marcus2020promoting}. On the other hand, the term `lifetime' could be understood to refer specifically to the active period of the AI system in operation~\cite{murakami2010lifespan}, while `lifecycle' could encompass a broader view of all phases from product design and development to decommissioning~\cite{hamon2024three}. In this case, however, it is unclear how accuracy, robustness and cybersecurity should be ensured beyond the operational phase (e.g., during development). 
\Art{2(8)} clarifies that these requirements do not have to be met during the test and development phase of the \HRAIS--unless the system is tested under real world conditions.
However, the use of the term 'lifecycle' might be interpreted to suggest that the requirements of \Art{15} should not only be assessed when the system is ready for deployment but also be considered during design process itself.

Second, it is unclear what `consistent' performance means and how it should be measured.
In the ML literature, a model's variability in performance over time is often measured using the variance of a metric such as accuracy or robustness~\cite{kilbertus2020fair, bechavod2019equal, rateike2022don}.
The variance of a metric over a time interval indicates its deviation from its mean within this interval. 
For instance, high variance in robustness indicates significant fluctuations in robustness levels between two points in time, whereas low variance indicates similar levels of robustness over time. 
A low variance could therefore be understood as a consistent performance.\footnote{Some also consider consistency as a metric itself, rather than as a property of a (robustness) metric~\cite{wei2020optimal}.} 
In practice, performance can vary due to factors, such as random initializations of weights or input data sampling.
These types of variations are unavoidable.
Defining level of variance considered `consistent' is challenging as it is dependent on the context.
Technical standards and guidelines by the EU Commission should clarify how to measure a consistent performance with respect to accuracy, robustness, and cybersecurity, and provide guidance on determining the required level of consistency.

 \subsection{Robustness \Art{15(4)}}\label{sec:challenges_robustness}
We now turn to 
challenges specific to \Art{15(4)}.
\Art{15(4)(i)} states that ``technical and organisational measures shall be taken'' to ensure that AI systems are ``as resilient as possible regarding errors, faults or
inconsistencies that may occur within the system or the environment''. \Art{15(4)(ii)} specifies that robustness can be achieved through technical redundancy
solutions, and \Art{15(4)(iii)} requires addressing feedback loops in online learning with possibly biased outputs.

\paragraph{Inconsistent Terminology.}
The term robustness is used inconsistently throughout the \EUAIAct.
\Artx\ 15(1) and (4) \EUAIAct\ refer to robustness, whereas the corresponding \EW{27} and \EW{75} both mention technical robustness. 
One could argue that technical robustness is synonymous with robustness. 
The term `technical robustness' in \EW{27} may be a remnant of the legislative process that built on the 2019 Ethics Guidelines for Trustworthy AI~\cite{aiiheg2019guidelines} developed by the AI IHEG, which introduced the principle of `technical robustness and safety'.  
These guidelines are explicitly referenced by \EW{27}.
Nevertheless, it remains unclear why \EW{75} also refers to `technical robustness'.
It could be that the wording in \EW{75} is borrowed from \EW{27}.
Alternatively, one could argue that robustness in \Artx\ 15(1) and (4) \EUAIAct\ is not limited to technical aspects, but additionally includes some form of non-technical robustness. 
The latter could refer to organizational measures that must be implemented to ensure robustness (\Art{15(4)(i)}).
Technical standards and guidelines by the EU Commission should clarify what aspects robustness encompasses.

\paragraph{Required Level of 
Robustness.} 
The \EUAIAct\ creates ambiguities regarding the required level of `robustness'.
\Art{15(1)} mandates that AI systems must achieve an ``appropriate level'' of robustness.
\Art{15(4)}, however, demands that AI systems shall be ``as resilient as possible'' to ``errors, faults, or inconsistencies'', suggesting 
a stricter requirement. 
This discrepancy initially appears ambiguous, as it is unclear whether \HRAIS\ must simply meet an appropriate standard of robustness or strive for the highest possible level. 
However, the ``appropriate'' level stated in \Art{15(1)} can be understood as a general principle, which is further specified by \Art{15(4)}. Therefore, appropriate with respect to robustness is to be understood as `as resilient as possible'. 

When determining the appropriate level of robustness of a specific \HRAIS, the intended purpose of the system and the generally acknowledged state of the art (SOTA) on AI and AI-related technologies must be taken into account (\Art{8(1)}).
\Art{9(4)} acknowledges that one of the objectives of the required risk management is to achieve an ``appropriate balance in the implementation of measures to fulfil'' requirements. 
\Art{9(5)} further acknowledges the permissibility of a residual risk, meaning that the measures adopted under the risk management system are not expected to eliminate all existing risks, but rather to maintain these residual risks at an `acceptable' level.
The risk management system is a continuous iterative process (\Art{9(1)}).
This means that the appropriate level of robustness of \HRAIS\ must be regularly determined and updated, taking into account its purpose and the SOTA while balancing it with other requirements.

\paragraph{Feedback Loops.}
\Art{15(4)(iii)} states that AI systems must be explicitly developed in such a way that they ``duly address'' feedback loops and ``eliminate or reduce'' the risks associated with them.
According to \EW{67}, feedback loops occur when the output of an AI system influences its input in future operations, an under
understanding that aligns with the concept as found in the ML literature.
Feedback loops are a well-studied problem manifesting in various forms~\cite{pagan2023classification}, with the most common issues being a distribution shift~\cite{perdomo2020performative} or a selection bias~\cite{kilbertus2020fair, lum2016predict}.
Importantly, in this context, the risk of `biased outputs' in feedback loops (\Art{15(4)(iii)}) is often studied in the literature on fairness in ML rather than in the literature on \emph{robustness} in ML, which traditionally constitute different research fields and communities~\citep{lee2021machine}.
\footnote{For example, whether there is a trade-off between \emph{robustness} and fairness, or if both pursue similar goals, remains an active discussion in the ML community~\citep{lee2021machine, xu2021robust, pruksachatkun2021does}.
}

An important aspect of \Art{15(4)(iii)} is that it applies specifically to AI systems that learn online. 
Online learning ML models iteratively learn from a sequence of data and continuously update their parameters over time~\cite{hoi2021online}.
This adaptiveness is reflected in \Art{3(1)} as a factual characteristic of an AI system.
The problem with feedback loops in online learning is that newly collected training data can become biased, e.g., due to selection bias, which occurs when the data collected is not representative of the overall population~\cite{zadrozny2004learning, liu2014robust}. This can distort model predictions and reinforce existing biases, ultimately impacting the model's accuracy and fairness~\cite{kilbertus2020fair, bechavod2019equal, rateike2022don}.
Offline models, in contrast, are trained on a fixed dataset all at once~\cite{hoi2021online}.
Offline models can also carry risks when feedback loops are present: The outputs of an ML model can induce a distribution shift through their interaction with the environment~\cite{d2020fairness, zhang2020fair, liu2018delayed}. Since an offline ML model is not updated, distribution shifts can influence their performance over time and possibly lead to fairness concerns~\cite{liu2018delayed}. 
Although \Art{15(4)} does not explicitly address feedback loops in offline systems, \HRAIS\ are not exempt from addressing them.
Since they can impact the model's accuracy, feedback loops in offline systems may still need to be addressed to comply with \Art{15(1)}.

 \subsection{Cybersecurity \Art{15(5)}}\label{sec:challenges_cybersecurity}

We now turn to legal challenges specific to \Art{15(5)}. \Artx\ 15(5)(i) \EUAIAct\ states that AI systems shall be resilient against attempts to ``alter their use, outputs, or performance by exploiting system vulnerabilities''. \Art{15(5)(ii)} specifies that technical solutions aiming to ensure resilience against such malicious attempts ``shall be appropriate to the relevant circumstances and the risks''. Finally, \Art{15(5)(iii)} mandates specific measures ``to prevent, detect, respond to, and control for attacks'' exploiting AI-specific vulnerabilities. 
This section examines the key aspects of compliance with \Art{15(5)}.
However, a mentioned above, providers have an additional pathway for demonstrating compliance with its cybersecurity requirements, namely a certification under the CSA~\cite{casarosa2022cybersecurity}.

\paragraph{Required Level of Cybersecurity.}

\Art{15(5)(ii)} mandates that technical solutions must be ``appropriate to the relevant circumstances and the risks'', 
but this needs further clarification.
The \EUAIAct\ specifically addresses only three kinds of risks: health, safety, and fundamental rights (\EW{1}). 
Risks associated with these aspects can be identified and managed through a risk management system that must be put into place as stipulated by \Art{9}.
Relevant circumstances are any known and foreseeable circumstances that may have an impact on cybersecurity.\footnote{See \Art{13(3)(b)(ii)} and Appendix~\ref{apx:relevant-circumstances}.}

Mandating a cybersecurity level that is `appropriate to the relevant circumstances' acknowledges that complex ML models generally cannot be expected to be fully resistant to all types of adversarial attacks.
This has two major reasons: First, it is impossible to anticipate all types of possible attacks. 
This is acknowledged by \Art{9(5)} which states that measures adopted under the risk management system are not expected to remove all existing risks.
Second, complete protection against a specific attack cannot be guaranteed, especially as adversaries continuously adapt their strategies to overcome possible defence mechanisms~\citep{xie2023defending, kumar2023certifying}. 
The appropriateness of a certain performance level must consider the intended purpose of the system and the generally acknowledged SOTA (see \Art{8(1)}).
The measures to ensure cybersecurity adopted are not expected to eliminate all existing risks, but the overall residual risk must be acceptable (see \Art{9(1) and (4)}).
Thus, when determining the appropriateness of technical solutions, all applicable requirements of the \EUAIAct\ must be balanced, while also mitigating risks to health, safety, and fundamental rights.

Lastly, while \HRAIS\ are expected to be `as resilient as possible' in terms of robustness, they need only to be `resilient' in terms of cybersecurity. Consequently, the wording of \Art{15} suggests that the robustness requirements are stricter. This may be due to the nature of unintentional causes—such as errors, faults, inconsistencies, or unexpected situations within the system or its operating environment—which are primarily within the provider's control and justify a higher duty of care. In contrast, attacks by unauthorized third parties are less controllable and therefore justify a (slightly) lower standard for the provider's duty regarding cybersecurity.

\paragraph{AI-specific Vulnerabilities.}
\Art{15(5)} differentiates between 'system vulnerabilities' (\Art{15(5)(i)}) and 'AI-specific vulnerabilities' (\Art{15(5)(iii)}). 
As the term vulnerability is not defined, 
we provide a working definition.
The United States' Common Vulnerabilities and Exposures (CVE) system defines vulnerability as ``[a]n instance of one or more weaknesses [...] that can be exploited, causing a negative impact to confidentiality, integrity, or availability''~\cite{cve_glossary}.
\Art{15(5)(iii)} provides a non-exhaustive list of components of an AI system that expose AI-specific vulnerabilities, such as training data, pre-trained components used in training, inputs, or the AI model.
However, there might be additional components of the AI system that may also harbour AI-specific vulnerabilities. 
The question is how to identify them.
We suggest performing a hypothetical test.
AI models play a central role in an AI system. If a vulnerability would be eliminated by replacing the AI model with a non-AI model, it should be deemed `AI-specific'. To define a non-AI model, 
we return to the definition of an AI system under the \EUAIAct.
It has been argued that the central characteristic of an AI system is its ability to infer from input to output~\cite{hacker2024comments}. 
This inference ability is typically performed by one or more AI models within an AI system. 
Therefore, non-AI models are all models lacking inference capability, such as rule-based decision-making systems that rely on predefined rules and logic defined by human experts.\footnote{Note that non-AI rule-based systems use human-defined rules, while rule-based ML models infer rules from data~\cite{naik2023machine, weiss1995rule}, qualifying as AI models. In a different context, the AI IHEG ethics guidelines~\cite{aiiheg2019guidelines} suggest fallback plans where AI systems  switch from a statistical (ML) approach to a rule-based or human-in-the-loop approach.}
Since AI-specific vulnerabilities relate to specific components of an AI system, we suggest viewing them as a subset of system vulnerabilities. 
To enhance clarity, technical standards should define both terms and mandate a process for identifying them.

\paragraph{Technical Solutions.}
\Art{15(5)(iii)} provides a non-exhaustive list of attacks and AI-specific vulnerabilities that must be addressed through technical solutions:
data poisoning, model poisoning, adversarial examples, model evasion, and confidentiality attacks, which are well-established in the ML literature.
These attacks aim to induce model failures~\cite{vassilev2024adversarial}:
\emph{Data poisoning} attacks manipulate training data~\citep{schwarzschild2021just}, \emph{model poisoning} attacks manipulate the trained ML model~\cite{zhang2022fldetector}, and \emph{model evasion} attacks 
manipulate test samples~\citep{biggio2013evasion}.
\emph{Confidentiality attacks}, typically explored in the field of privacy in ML, refer to attempts to extract information about the training data or the model itself~\citep{rigaki2023survey}. 

In addition to these attacks, 
 \Art{15(5)(iii)} lists `model flaws' as an AI-specific vulnerability.
 This is a vague legal term and lacks an established counterpart in the ML literature. 
 In software contexts, the word \emph{flaw} often refers to so-called \emph{bugs}, which are typically the result of human errors in the coding process~\cite{kumar2023, nissenbaum1996accountability}. 
 However, the term model flaw follows the list of attacks outlined above, which
 are instead designed to exploit the default properties of a properly functioning ML model, and are not directly the results of errors in the coding process.
 Thus, it is unclear what model flaw refers to in this context, and whether technical solutions are only expected to address traditional bugs or coding errors, or whether 
 they should address other ways of exploiting AI-specific vulnerabilities.
Given that the term is situated within the cybersecurity requirements for AI system outlined in \Art{15(5)}, we argue that the term model flaws should be interpreted as flaws that enable the exploitation of AI-specific vulnerabilities.
Technical standards and guidelines by the EU Commission should define model flaws more clearly and provide guidelines for technical solutions to address these model flaws.
This should take into account the arms race between attacker and defender in the realm of adversarial robustness, in which both parties are continuously adapting their strategies to outmanoeuvre each other~\cite{chen2017adversarial}.
This makes it infeasible to anticipate and counter all potential attacks that target AI-specific vulnerabilities.

\paragraph{Organizational Measures.}
Numerous EU regulations related to cybersecurity (see e.g., \Artx\ 32 General Data Protection Regulation\footnote{EU Regulation 2016/679, OJ L 119, 4.5.2016.}, Art. 21 NIS 2 Directive\footnote{EU Directive 2022/2555, OJ L 333/80.}) explicitly mandate both technical and organizational measures to ensure cybersecurity. 
In the \EUAIAct,
organizational measures are only mandated for the robustness of \HRAIS\ in \Art{15(4)}, but not for cybersecurity (\Art{15(5)}).
Rather, \Art{15(5)} \EUAIAct\ only focuses on technical solutions for providers of \HRAIS.
The omission of organizational measures in \Art{15(5)} has been criticized in the literature accompanying the legislative process of the \EUAIAct~\cite{biasin2023new}.
It is unclear whether providers are still implicitly required to implement organizational measures (in accordance with other EU regulations), as these measures might be inherently included in the concept of cybersecurity, or if they are not mandatory.
However, this ambiguity for \emph{providers} of \HRAIS\ (\Art{15}) is mitigated by the fact that \emph{deployers} of \HRAIS\ are required to implement both organizational and technical measures to ensure the proper use of the system in accordance with the instructions for use (\Art{26(1)}). These instructions include the cybersecurity measures put in place.

\section{Requirements for General-Purpose AI Models With Systemic Risk}\label{sec:gpai}

In the previous section, we examined \HRAIS\ requirements. To further elucidate them,
we study \GPAIMSSR, highlighting similarities and differences. 
The \EUAIAct\ establishes legal requirements for \GPAIMS, such as multimodal large language models~\cite{openai2023gpt, team2023gemini},
which can perform tasks beyond their original training objective~\citep{gutierrez2023proposal}.
\GPAIM\ can be stand-alone or embedded in an \HRAIS, with the latter requiring compliance with both \GPAIM\ and \HRAIS\ requirements.
The \EUAIAct\ distinguishes between \GPAIM\ with systemic risks and those without.
\Art{3(65)} defines `systemic risk' as the risk that is specific to the high-impact capabilities of \GPAIMS\ that have a ``significant impact'' on the market, public health, safety, security, fundamental rights, or society.\footnote{A systemic risk is presumed when the cumulative computation during training exceeds \(10^{25}\) Floating-Point Operations Per Second (FLOPS). \GPAIMS\ with fewer FLOPS may still be classified as posing a systemic risk under \Art{51(1)}. 
There is an ongoing debate over this threshold and if a model's complexity truly reflects its risk level~\cite{hacker2024comments, novelli2024generative, pehlivan2024eu, kutterer2023regulating}. Thresholds and criteria can be modified by the EU Commission, see Appendix~\ref{apx:hrais-gpaim}
}
\GPAIMS\ without systemic risks are exempt from robustness and cybersecurity obligations (\Art{53} ff.).

\paragraph{Cybersecurity Requirements.} 
\Art{55(1)(d)} mandates ``an adequate level of cybersecurity protection'' for \GPAIMSSR.
\EW{115} further details this cybersecurity requirement.
It mandates cybersecurity protection against ``malicious use or attacks'' and lists specific adversarial threats, such as ``accidental model leakage, unauthorised releases, circumvention of safety
measures'', ``cyberattacks'', or ``model theft''. 
Notably, several of these threats have direct counterparts in the ML literature on \emph{adversarial robustness} and \emph{privacy} for large generative models, such as the circumvention of safety measures (jailbreaking) or model theft~\cite{yao2024survey, li2023multi, wang2023survey}.
Although \Art{55(1)(d)} does not define the term `cyberattacks', we infer that it includes the attacks exploiting AI-specific vulnerabilities mentioned in \Art{15} (see Section~\ref{sec:challenges_cybersecurity}).
These attacks are studied in the field of \emph{adversarial robustness} and 
can also affect \GPAIMS~\cite{qiang2024learning, das2024exposing, yan2024backdooring, schwinn2024soft, vitorino2024adversarial}---even though specific ML techniques may be necessary to address \GPAIM-specific challenges. 
This relation underscores that the concepts and problems explored under \emph{adversarial robustness} are reflected in the term `cybersecurity' as used in \Art{55(1)(d)}.
To ensure the cybersecurity of a \GPAIMSSR, providers must conduct and document internal and/or external adversarial testing of the model, such as red teaming.\footnote{In this context, red teaming refers to stress testing AI models by simulating adversarial attacks~\cite{feffer2024red}, such as linguistic or semantic attacks against LLMs~\cite{shi2024red}, whereas traditional cybersecurity red teaming focuses on assessing entire systems or networks.~\cite{teichmann2023overview}}

While the \EUAIAct\ mandates robustness requirements for \HRAIS, we observe that it does not impose an explicit equivalent legal requirement for \GPAIMS, regardless of whether they present a systemic risk or not.
Specifically, neither \Art{55} nor \EW{155} address unintentional causes for deviations from consistent performance. 
In Section~\ref{sec:challenges_robustness}, we stated that \emph{non-adversarial robustness} is reflected in the term robustness in \Art{15}. 
Consequently, \GPAIMS, which are not required to fulfil any robustness requirement, are not mandated to be resilient against performance issues, such as data distribution shifts or noisy data.
The \EUAIAct\ itself does not provide an explanation for the omission of a robustness requirement.
It may stem from the complexity of political negotiations regarding the \EUAIAct, particularly regarding \GPAIMS, which were not addressed in the initial draft of the regulation but gathered widespread media attention during the legislative procedure.
However, evidence from ML research suggests that \emph{non-adversarial robustness} is also relevant for \GPAIMS~\cite{yuan2023revisiting, chen2022foundational}.

\paragraph{Required Level of Cybersecurity.}
\Art{55(1)(d)} mandates an `adequate' level of cybersecurity protection for \GPAIMSSR. This requirement contrasts with the `appropriate' level of cybersecurity mandated for \HRAIS\ under \Art{15(1)}.
The use of these two different terms raises questions about whether both \HRAIS\ and \GPAIMS\ should achieve the same level of cybersecurity or to what extent their required levels might differ.
On the one hand, `adequate' and `appropriate' could imply different levels of cybersecurity. The Cambridge Dictionary defines the term `adequate' as ``enough or satisfactory for a particular purpose''~\cite{CambridgeAdequate} and `appropriate' as ``suitable or right for a particular situation or occasion''~\cite{CambridgeAppropriate}.
Accordingly, something is `adequate' if it exceeds a minimum threshold that is good enough, while something is `appropriate' if it meets a specific (right) level above that minimum.
GPAI models can perform a wide variety of tasks in different contexts and thus be prone to a variety of different intentional causes of harm,  
 making it difficult to identify and mitigate their specific cybersecurity risks.
For this reason, it may be reasonable to only mandate an `adequate', i.e., minimum level of cybersecurity.
\HRAIS, independently of whether they contain an \GPAIM\ as a component, can be thought of as operating in a more specific contexts, potentially allowing an easier and more precise assessment of cybersecurity risks and thus a more stringent appropriate level of cybersecurity protection.
On the other hand, `adequate' and `appropriate' could refer to the same level of cybersecurity.
\EW{115} 
states that ``adequate technical and established solutions'' must be ``appropriate to the relevant circumstances and the risks''. The simultaneous use of both terms in a single sentence, intended to guide the interpretation of  \Art{55(1)(d)}, suggests that they might be intended as synonymous. This is corroborated by the observation that many official language versions of the \EUAIAct\ use a single term for both ``adequate'' and ``appropriate'' in \Art{15(1)} and \Art{55(1)(d)}.\footnote{Such as FR ``approprié'', ES ``adecuado'', GER ``angemessen'', IT ``adeguato''.} 
To resolve this ambiguity, technical standards should clarify the required level of cybersecurity for \GPAIMSSR.

\section{Summary and Outlook}\label{sec:discussion_outlook}
We identified several legal challenges and potential limitations in implementing robustness and cybersecurity requirements for \HRAIS\ under \Art{15(4) and (5)}. 
We also examined \GPAIMSSR, which face cybersecurity but not robustness requirements, and identified additional legal challenges.
We proposed a simple explanatory model that maps \emph{non-adversarial robustness} in the ML literature to the term `robustness' used in \Art{15(1) and (4)}, and that maps \emph{adversarial robustness} to the term `cybersecurity' used in \Art\ 15(1) and (5) \EUAIAct.
However, 
both `robustness' and `cybersecurity' can refer also to other concepts both within the domain of ML and beyond.
Comparing the provisions for \HRAIS\ to those for \GPAIMSSR, we argued that \emph{adversarial robustness} maps to the term `cybersecurity' used in \Art{55(1)(d)}. However, we were not able to find an explicit equivalent legal requirement for \emph{non-adversarial robustness} in the provisions regulating \GPAIMSSR\ models.

Our analysis highlights the need for clearer specifications of these provisions through harmonized standards, guidelines by the EU Commission or the benchmark and measurement methodologies foreseen for robustness and cybersecurity in \Art{15(2)}. These would help define technical requirements and establish evaluation criteria for AI systems.
Specifically, we suggest that technical standards and guidelines by the EU Commission should focus on: i) identify the technical requirements associated with vague legal terms; ii) defining the required level of `robustness' and `cybersecurity' and other concepts such as `consistency'; iii) defining the requirements for evaluating and assessing AI systems and its components; and iv) pay attention to some aspects that are not explicitly regulated, such as feedback loops in offline systems in \Art{15(4)} or   organizational measures to ensure `cybersecurity' in \Art{15(5)}.

However, while standards and guidelines can ensure compatibility and practical integration of regulatory frameworks, they can struggle to keep pace with rapid technological advancements. 
This can lead to outdated versions that do not fully address emerging technologies or novel applications.
Our analysis is not without limitations. Due to the novelty of the \EUAIAct, we lack empirical data to support claims about the challenges in implementing its requirements. While we focus on identifying these challenges, proposing specific definitions, processes, metrics, or thresholds is left for future work.
Future research should focus on non-adversarial robustness for \GPAIMSSR,
and explore legal intersections with frameworks like the Medical Device Regulation~\cite{biasin2023new, nolte2024new}. Additionally, the focus on models in ML research versus entire AI systems in the \EUAIAct\ underscores the need for interdisciplinary work. Within the ML domain, future work should explore the impact of accuracy metrics on robustness, potential `robustness hacking', and methods to measure and ensuring long-term performance consistency in the presence of feedback loops.

 \begin{acks}
Thank you to Tommaso Fia, and Sebastian Bordt for helpful feedback and comments.
Special thank you to Marie-Sophie Müller for a thorough review of our text.

Michèle Finck is a members of the Machine Learning Cluster of Excellence, funded by the Deutsche Forschungsgemeinschaft under 
Germany’s Excellence Strategy – EXC number 2064/1 – Project number 
390727645.
Michèle Finck and Henrik Nolte thank the Carl Zeiss Foundation for funding support.
Miriam Rateike is grateful for the generous funding support by the 2023 Google PhD Fellowship in Machine Learning.
\end{acks}

\bibliographystyle{ACM-Reference-Format}
\bibliography{main}

\clearpage
\section*{Appendix}
\appendix
\section{Legal Terminology}\label{apx:legal}

The \EUAIAct\ is formally structured into recitals (\EWx), articles (\Artx), and annexes. Recitals are legally non-binding and outline the rationale behind the articles, articles delineate specific binding obligations, and the annexes provide additional details and specifications to support the articles~\citep{klimas15law}.

\section{Notions of High-Risk AI Systems and General Purpose Models provided in \EUAIAct }\label{apx:hrais-gpaim}

\paragraph{AI System.} The \EUAIAct\ defines an AI system as ``a machine-based\footnote{Thereby,
machine-based ``refers to the fact that AI systems run on machines'' (\EW{12}).} system that is designed to operate with varying levels
of autonomy and that may exhibit adaptiveness after deployment, and that, for explicit or
implicit objectives, infers, from the input it receives, how to generate outputs such as
predictions, content, recommendations, or decisions that can influence physical or virtual
environments'' (\Art{3(1)}). 
Thereby, \EW{12}, suggests that notion of `AI system' ``should be clearly defined and should be
closely aligned with the work of international organisations working on AI to ensure legal
certainty, facilitate international convergence and wide acceptance, while providing the
flexibility to accommodate the rapid technological developments in this field''. Importantly, ``the definition should be based on key characteristics of AI systems that distinguish it from
simpler traditional software systems or programming approaches and should not cover
systems that are based on the rules defined solely by natural persons to automatically
execute operations'' (\EW{12}). 
Thereby a ``key characteristic of AI systems is their capability to infer'', which refers to ``the process of obtaining the outputs, such as predictions,
content, recommendations, or decisions, which can influence physical and virtual
environments, and to a capability of AI systems to derive models or algorithms, or both,
from inputs or data'' (\EW{12}). Examples for techniques that enable inference 
``include machine learning approaches that learn from data how to achieve certain
objectives, and logic- and knowledge-based approaches that infer from encoded knowledge
or symbolic representation of the task to be solved'' (\EW{12}).

\paragraph{High-risk AI Systems (HRAIS).} The \EUAIAct\ classifies AI systems into different risk groups.
Risk refers thereby to ``the combination of the probability of an occurrence of harm and the severity
of that harm'' (\Art{3(2)}).
An AI system is considered high-risk, if it is ``intended to be used as a safety component of a product, or the AI
system is itself a product'' and ``is required to undergo a third-party conformity
assessment, with a view to the placing on the market or the putting into service of that product'' covered by the Union harmonisation legislation listed in
Annex I (\Art{6(1)}).
Annex I \EUAIAct\ provides a list of 20 EU harmonisation legislation. 
For example, the Machinery Directive (Directive 2006/42/EC and amending  Directive 95/16/EC), the directive on the safety of toys (Directive 2009/48/EC), or the directive concerning agricultural and forestry
vehicles (Regulation (EU) No 167/2013).
This means that for example, an AI system that is used in a product or as a product itself considered as a toy falling under Directive 2009/48/EC, would be considered as a high-risk AI system.

An AI systems is also considered high-risk, if it operates in types or use-cases enlisted in
Annex III (\Art{6(2)}) and poses a significant risk of harm to the health,
safety or fundamental rights of natural person following (\Art{6(3)}).
Annex III enlists eight types or use-cases of AI systems: biometric applications (e.g., remote biometric identification systems excluding biometric verification, biometric categorisation, and emotion recognition); critical infrastructure (e.g., ``critical digital infrastructure, road traffic, or in the supply of
water, gas, heating or electricity''); education and vocational training; employment, workers management and access to self-employment; essential private and public services; law enforcement; migration, asylum and border control management; and administration of justice and democratic processes. 
Thereby, the Commission shall ``provide [...] a comprehensive list of practical examples of use cases
of AI systems that are high-risk and not high-risk'' (\Art{6(5)}).
 \EW{52} suggests that ``it is appropriate to classify
them [i.e., AI systems] as high-risk if, in light of their intended purpose, they pose a high risk of harm to the
health and safety or the fundamental rights of persons, taking into account both the severity
of the possible harm and its probability of occurrence and they are used in a number of
specifically pre-defined areas''. 
Thereby the methodology and criteria for the identification of high-risk should be able to be adopted in order to 
account for ``the rapid pace of technological
development, as well as the potential changes in the use of AI systems'' (\EW{52}).
For example, an AI system used for remote biometric identification is considered high-risk, unless it does not pose a significant risk. Remote identification refers to comparing biometric data of that individual to stored biometric data of
individuals in a reference database (\EW{15}), where a remote biometric identification system should be understood as a ``AI system intended for the identification of natural persons
without their active involvement, typically at a distance'' (\EW{17}).

\paragraph{General-purpose AI Models (GPAIM).}
A general-purpose AI model (\GPAIM) refers to an ``AI model [...] that displays significant
generality and is capable of competently performing a wide range of distinct tasks
regardless of the way the model is placed on the market and that can be integrated into a
variety of downstream systems or applications'' (\Art{3(63)}). 
The \EUAIAct, however, explicitly excludes from its regulation AI models that may fall under this definition but are ``used for
research, development or prototyping activities before they are placed on the market'' (\Art{3(63)}).
\EW{97} suggest that the term general-purpose AI model ``should be clearly defined and set apart from the
notion of AI systems to enable legal certainty'' taking into account the ``key
functional characteristics of a general-purpose AI model, in particular the generality and
the capability to competently perform a wide range of distinct tasks''. 
Thereby, GPAIMs ``may
be placed on the market in various ways, including through libraries, application
programming interfaces (APIs), as direct download, or as physical copy'', and ``may be further modified or fine-tuned into new models'' (\EW{97}).
Examples for GPAIMs, are large generative AI models that ``allow for flexible generation of content, such as in the form of text, audio, images
or video, that can readily accommodate a wide range of distinctive tasks'' (\EW{99}).
This is the case for many multimodal large language models, such as GPT-4 Omni (GPT-4o)~\footnote{\url{https://openai.com/gpt-4o-contributions/}}, or Gemini~\footnote{\url{https://gemini.google.com}}.

\paragraph{\GPAIM\ with Systemic Risk.} 
The systemic risk of a \GPAIM\ is understood as ``a risk that is specific to the high-impact capabilities of
general-purpose AI models, having a significant impact on the Union market due to their
reach, or due to actual or reasonably foreseeable negative effects on public health, safety,
public security, fundamental rights, or the society as a whole, that can be propagated at
scale across the value chain'' (\Art{3(65)}). 
\EW{110} provides as a non-exhaustive list of examples for systemic risks:
``any actual or reasonably foreseeable negative effects in relation to major accidents,
disruptions of critical sectors and serious consequences to public health and safety; any
actual or reasonably foreseeable negative effects on democratic processes, public and
economic security; the dissemination of illegal, false, or discriminatory content''. Thereby ``[s]ystemic
risks should be understood to increase with model capabilities and model reach, can arise
along the entire lifecycle of the model, and are influenced by conditions of misuse, model
reliability, model fairness and model security, the level of autonomy of the model, its
access to tools, novel or combined modalities, release and distribution strategies, the
potential to remove guardrails and other factors'' (\EW{110}). 

According to \Art{51(1)(a)}, a GPAIM is considered posing a
systemic risk, if ``it has high impact capabilities evaluated on the basis of appropriate technical tools
and methodologies, including indicators and benchmarks''. 
High-impact capabilities are understood as ``capabilities that match or exceed the capabilities recorded
in the most advanced general-purpose AI models'' (\Art{3(64)}).
Thereby high impact capabilities of a GPAIM are to be assumed, ``when the cumulative amount of computation used for its training
measured in floating point operations is greater than $10^{25}$'' (\Art{51(2)}).
\EW{111} states that ``according to the state of
the art at the time of entry into force of this Regulation, the cumulative amount of
computation used for the training of the general-purpose AI model measured in floating
point operations is one of the relevant approximations for model capabilities''.
According to \Art{51(1)(b)} a GPAIM is also considered to pose a systemic risk if---``based on a decision of the Commission''---it is considered having capabilities or an impact
equivalent to those set out in \Art{51(1)(a)} according to the criteria set out in Annex XIII.
The seven criteria enlisted in Annex XIII include the number of model parameters, the quality or size of the data set,  the amount of computation used for training the model, the input and output modalities of the model, the benchmarks and evaluations of capabilities of the model, whether it has a high impact on the internal market due to its reach, and the number of registered end-users.

Importantly, according to \Art{51(3)}, the EU Commission shall be able to amend the
thresholds in \Artx\ 51(1) and (2) \EUAIAct\ and add ``benchmarks
and indicators in light of evolving technological developments, such as algorithmic
improvements or increased hardware efficiency, when necessary, for these thresholds to
reflect the state of the art'' (see also \EW{111}).

\section{Lifecycle in \Art{15(1)}}\label{apx:lifecycle}

The exact timeframe during which consistent performance must be 
ensured following \Art{15(1)} is unclear. Particularly, the term `lifecycle' is not defined, leaving open whether it differs from the term `lifetime' used in \Art{12(1)} and \EW{71}.
It is crucial to clarify the exact timeframe during which consistent performance must be maintained.
While `lifecycle' and `lifetime' could initially be interpreted as synonyms~\cite{marcus2020promoting}, `lifetime' might refer specifically to the active operational period of the AI system~\cite{murakami2010lifespan}, whereas `lifecycle' could encompass a broader view of all phases from product design and development to decommissioning~\cite{hamon2024three}. If this broader interpretation of `lifecycle' is intended, it raises questions about how accuracy, robustness, and cybersecurity should be ensured beyond the operational phase (e.g., during development), and why this would be necessary when there are no immediate risks to health, safety, and fundamental rights. One explanation for using the term 'lifecycle' would be that the EU legislator intended to emphasize that the requirements of \Art{15} should not only be assessed when the system is ready for deployment but also during the design process. Accordingly, technical standards should define both terms.

\section{Relevant Circumstance \Art{15(5)(ii)}}\label{apx:relevant-circumstances}

The term `relevant circumstance' is not defined in the \EUAIAct\ and therefore requires interpretation.
On the one hand, one could argue that the term only refers to circumstances 
that are ``important'' for a ``particular purpose'' or context~\cite{CambridgeRelevant}.
On the other hand, the meaning of the term can also result from a comparison with other provisions of the \EUAIAct\ such as 
\Art{13(3)(b)(ii)}, which suggests a different understanding. 
The provision demands that the instructions for the use of AI systems shall contain ``any known and foreseeable circumstances'' that may have an impact on cybersecurity. 
This speaks for a broader understanding of relevance, which only excludes unknown and unforeseeable circumstances.
Given this ambiguity, standards should elaborate on how to determine relevant circumstances.

\section{Robustness-Accuracy Trade-Off}\label{apx:robustness-accuracy-trade-off}
The ML literature has found that robustness and accuracy of ML models can in some scenarios be empirically and theoretically mutually inhibiting~\cite{zhang2019theoretically, li2024triangular, yang2020closer, raghunathan2020understanding, rade2022reducing}, and this relationship remains an active area of research~\cite{pang2022robustness}. 
As stated in Section~\ref{sec:general_challenges}, an AI system may also incorporate other technical components (beyond models) that must be robust and may impact the overall system's accuracy.
The \EUAIAct\ acknowledges these trade-offs but does not offer specific guidelines on how to achieve this balance. 
It requires providers of \HRAIS\ to ensure an appropriate level of both accuracy and robustness, and to find “an appropriate balance in implementing the measures to fulfil” the \EUAIAct\ requirements (\Art{9(4)}). Additionally, the technical documentation must include ``decisions about any possible trade-off made regarding the technical solutions adopted to comply with the requirements'' of the \EUAIAct\ (Annex IV Nr. 2 lit. b \EUAIAct). Standards may offer guidance on the processes and metrics to use but will not prescribe a specific balance, such as a 40-60 ratio. As a result, providers of \HRAIS\ will ultimately need to navigate these complex trade-offs themselves, adjusting individual model parameters to find an appropriate balance and justify and document their decisions.

\end{document}